\documentclass[10pt,twocolumn,letterpaper]{article}

\usepackage{iccv}
\usepackage{times}
\usepackage{epsfig}
\usepackage{graphicx}
\usepackage{amsmath}
\usepackage{amssymb}
\usepackage[normalem]{ulem}
\usepackage{multirow}
\usepackage{booktabs}
\usepackage{wrapfig}
\usepackage{pifont}
\usepackage{arydshln}
\usepackage{tabularray}
\usepackage{iccv}
\usepackage{times}
\usepackage{epsfig}
\usepackage{graphicx}
\usepackage{amsmath}
\usepackage{amssymb}
\usepackage[accsupp]{axessibility} 

\usepackage[pagebackref=true,breaklinks=true,letterpaper=true,colorlinks,bookmarks=false]{hyperref}

\iccvfinalcopy

\def\iccvPaperID{9192} 

\ificcvfinal\fi

\begin{document}

\title{Diffusion-based Image Translation with Label Guidance for Domain Adaptive Semantic Segmentation}

\author{Duo Peng$^{1}$\quad Ping Hu$^{2}$\quad Qiuhong Ke$^{3}$\quad Jun Liu$^{1,}$\thanks{Corresponding Author}\\
$^{1}$Singapore University of Technology and Design\quad $^{2}$Boston University\\
$^{3}$Monash University\\
{\tt\small duo\_peng@mymail.sutd.edu.sg}, {\tt\small pinghu@bu.edu}, {\tt\small qiuhong.ke@monash.edu}, {\tt\small jun\_liu@sutd.edu.sg}
}

\maketitle
\ificcvfinal\thispagestyle{empty}\fi

\begin{abstract}
   Translating images from a source domain to a target domain for learning target models is one of the most common strategies in domain adaptive semantic segmentation (DASS). However, existing methods still struggle to preserve semantically-consistent local details between the original and translated images. In this work, we present an innovative approach that addresses this challenge by using source-domain labels as explicit guidance during image translation. Concretely, we formulate cross-domain image translation as a denoising diffusion process and utilize a novel Semantic Gradient Guidance (SGG) method to constrain the translation process, conditioning it on the pixel-wise source labels. Additionally, a Progressive Translation Learning (PTL) strategy is devised to enable the SGG method to work reliably across domains with large gaps. Extensive experiments demonstrate the superiority of our approach over state-of-the-art methods.
\end{abstract}

\begin{figure}[tp]
\centering{}
 \includegraphics[scale=0.40]{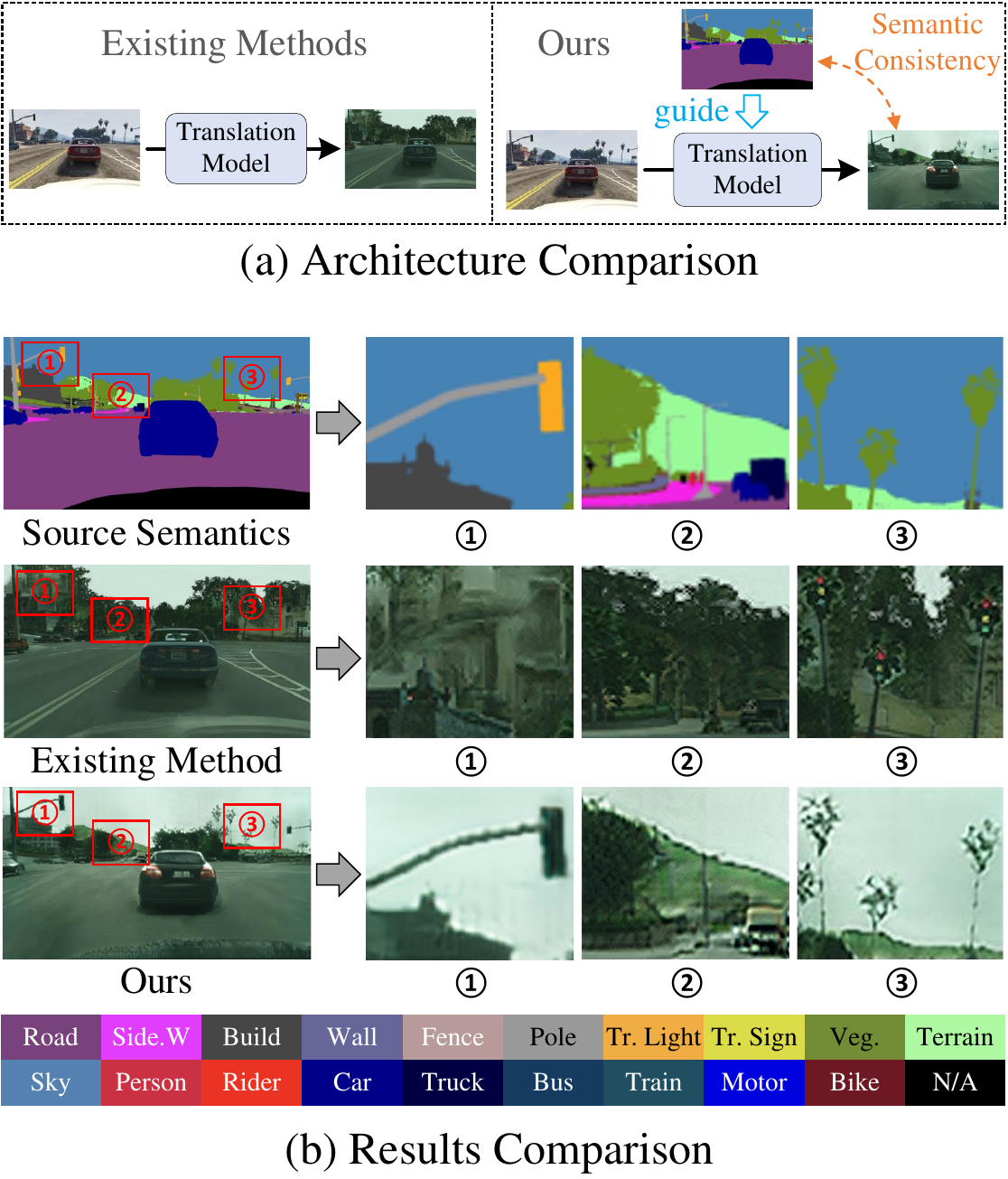} 
 \caption{\textbf{(a)} Architecture comparison between different image translation methods for DASS. Existing methods mainly translate images based on the input image. Our method further introduces the source semantic label to guide the translation process, making the translated content semantically consistent with the source label (as well as the source image).
\textbf{(b)} Results comparison in detail. First row: the source semantic label. Second row: the translated image of an existing method \cite{gao2021addressing}, which is a state-of-the-art image translation method based on GAN. Third row: the translated image of our method. The translated image of the existing method shows unsatisfactory local details (e.g., in case \ding{172} of the second row, the traffic light and its surrounding sky are translated into building). In contrast, our method preserves details in a fine-grained manner. 
}
\label{fig_abstract}
\end{figure}

\section{Introduction}

Semantic segmentation is a fundamental computer vision task that aims to assign a semantic category to each pixel in an image.
In the past decade, fully supervised methods \cite{long2015fully,ronneberger2015u,huang2019ccnet,zheng2015conditional} have achieved remarkable progress in semantic segmentation. 
However, the progress is at the cost of a large number of dense pixel-level annotations, which are significantly labor-intensive and time-consuming to obtain \cite{chen2017deeplab,long2015fully}.
Facing this problem, one alternative is to leverage synthetic datasets, e.g., GTA5 \cite{richter2016playing} or SYNTHIA \cite{ros2016synthia}, where the simulated images and corresponding pixel-level labels can be generated by the computer engine freely. 
Although the annotation cost is much cheaper than manual labeling, segmentation models trained with such synthetic datasets often do not work well on real images due to the large domain discrepancy.

The task of domain adaptive semantic segmentation (DASS) aims to address the domain discrepancy problem by transferring knowledge learned from a source domain to a target one with labels provided for the source domain only.
It is a very challenging problem as there can be a large discrepancy (e.g., between synthetic and real domains), resulting in great difficulties in knowledge transfer.
In DASS, one type of methods \cite{hoffman2016fcns,chen2017no,sankaranarayanan2017unsupervised,chen2018road,zhang2018fully,du2019ssf,luo2019significance,tsai2019domain,luo2019taking,sun2019not,wang2020differential} focuses on aligning the data distributions from different domains in the feature space, which obtains promising performance,
while another type of approaches \cite{zou2018unsupervised,zou2019confidence,li2020content,zhang2021prototypical,lee2022bi,li2022class} aims to generate pseudo labels for unlabeled target images, also making great achievements.
Besides above approaches, image translation \cite{zhu2017unpaired} is also an effective and important way to handle the DASS problem. 
It focuses on translating the labeled source images into the target domain, then using the translated images and source labels to train a target-domain segmentation model.
In the past few years, lots of DASS studies \cite{murez2018image,hoffman2018cycada,chen2019crdoco,li2019bidirectional,gao2021addressing} have been conducted based on image translation. This is because image translation allows for easy visual inspection of the adaptation method's results, and the translated images can be saved as a dataset for the future training of any model, which is convenient. 
Existing image translation methods mainly adopt generative adversarial networks (GANs) \cite{NIPS2014_5ca3e9b1} to translate images across a large domain gap.
However, due to the intractable adversarial training, GAN models can be notoriously hard to train and prone to show unstable performance \cite{saxena2021generative,miyato2018spectral}.
Furthermore, many studies \cite{matsunaga2022fine,dhariwal2021diffusion,nichol2021improved} have also shown that GAN-based methods struggle with preserving local details, leading to semantic inconsistencies between the translated images and corresponding labels, as shown in the second row of Fig. \ref{fig_abstract} (b).
Training the segmentation model on such mismatched image and label pairs will cause sub-optimal domain adaptation performance.
Facing these inherent limitations, the development of GAN-based image translation methods is increasingly encountering a bottleneck. 
Yet, there is currently little research exploring alternative approaches to improve image translation for DASS beyond GANs.

Denoising Diffusion Probabilistic Models (DDPMs), also known as diffusion models, have recently emerged as a promising alternative to GANs for various tasks, such as image generation \cite{dhariwal2021diffusion,vahdat2021score}, image restoration \cite{welker2022driftrec,yin2022diffgar}, and image editing \cite{ackermann2022high,meng2021sdedit}, etc. 
Diffusion models have been shown to have a more stable training process \cite{dhariwal2021diffusion,nichol2021improved,ho2020denoising}.
Inspired by its strong capability, we propose to exploit the diffusion model for image translation in DASS.
Specifically, based on the diffusion technique, we propose a label-guided image translation framework to preserve local details.
We observe that existing image translation methods \cite{hoffman2018cycada,murez2018image,chen2019crdoco,li2019bidirectional,li2020simplified} mainly focus on training a translation model with image data alone, while neglecting the utilization of source-domain labels during translation, as shown in Fig. \ref{fig_abstract} (a).
Since source labels can explicitly indicate the semantic category of each pixel, introducing them to guide the translation process should improve the capability of the translation model to preserve details.
A straightforward idea of incorporating the source labels is to directly train a conditional image translation model, 
where the translated results are semantically conditioned on the pixel-wise source labels.
However, it is non-trivial since the source labels and target images are not paired, which cannot support the standard conditional training.
Recent advances \cite{dhariwal2021diffusion,nichol2021improved,rombach2022high} have shown that a pre-trained unconditional diffusion model can become conditional with the help of the gradient guidance \cite{dhariwal2021diffusion}. 
The gradient guidance method can guide the pre-trained unconditional diffusion model to provide desired results by directly affecting the inference process (without any training).
In light of this, we propose to first train an unconditional diffusion-based image translation model, and then apply gradient guidance to the image translation process, making it conditioned on source labels.
However, we still face two challenges: (1) Traditional gradient guidance methods generally focus on guiding the diffusion model based on image-level labels (i.e., classification labels), while the DASS problem requires the image translation to be conditioned on pixel-level labels (i.e., segmentation labels).
(2) The gradient guidance methods typically work within a single domain, whereas the DASS task requires it to guide the image translation across different domains.
To address the first challenge, we propose a novel Semantic Gradient Guidance method (SGG), which can precisely guide the image translation based on fine-grained segmentation labels.
To tackle the second challenge, we carefully design a Progressive Translation Learning strategy (PTL), which is proposed to drive the SGG method to reliably work across a large domain gap.
With the help of SGG and PTL, our framework effectively handles the image translation for DASS in a fine-grained manner, which is shown in the third row of Fig. \ref{fig_abstract} (b).

In summary, our contributions are three-fold.
(i) We propose a novel diffusion-based image translation framework with the guidance of pixel-wise labels, which achieves image translation with fine-grained semantic preservation. To the best of our knowledge, this is the first study to exploit the diffusion model for DASS. (ii) We devise a Semantic Gradient Guidance (SGG) scheme accompanied with a Progressive Translation Learning (PTL) strategy to guide the translation process across a large domain gap in DASS. (iii) Our method achieves the new state-of-the-art performance on two widely used DASS settings. Remarkably, our method outperforms existing GAN-based image translation methods significantly on various backbones and settings.


\section{Related Work}

\textbf{Domain adaptive semantic segmentation (DASS)} is a task that aims to improve the semantic segmentation model's adaptation performance in new target scenarios without access to the target labels.
Current DASS methods \cite{hoffman2016fcns,sankaranarayanan2017unsupervised,li2020content,cheng2021dual,hoffman2018cycada} have pursued three major directions to bridge the domain gap.
The first direction aims to align data distributions in the latent feature space.
With the advent of adversarial learning \cite{ganin2016domain}, many methods \cite{hoffman2016fcns,chen2017no,pan2020unsupervised,sankaranarayanan2017unsupervised,chen2018road,lai2022decouplenet,zhang2018fully,du2019ssf,luo2019significance,tsai2019domain,luo2019taking,sun2019not,wang2020differential} adopt the adversarial learning to align feature distributions by fooling a domain discriminator.
However, the model based on adversarial learning can be hard to train \cite{zhang2022confidence,saxena2021generative,miyato2018spectral}, as it requires carefully balancing the capacity between the feature encoder (i.e., generator) and the domain discriminator, which is intractable. The second direction considers exploring the network's potential of self-training on the target domain \cite{zhang2021prototypical,ma2021coarse,li2020content,lee2022bi,wang2021uncertainty,li2022class,zou2018unsupervised,zou2019confidence}. 
Self-training methods generally use the source-trained segmentation model to predict segmentation results for target-domain images. 
They typically select the results with high confidence scores as pseudo labels to fine-tune the segmentation model. 
These works mainly focus on how to decide the threshold of confidence scores for selecting highly reliable pseudo labels. 
However, self-training is likely to introduce inevitable label noise \cite{li2019bidirectional,huo2022domain,yang2020label}, as we cannot fully guarantee that the selected pseudo labels are correct especially when the domain gap is large. 
The third direction is inspired by the image translation technique \cite{zhu2017unpaired}, aiming to transfer the style of source labeled images to that of the target domain \cite{cheng2021dual,murez2018image,hoffman2018cycada,chen2019crdoco,li2019bidirectional,gao2021addressing}. 
Inspired by the advances of the generative adversarial networks (GANs), they mostly adopt GANs as image translation models.
While remarkable progress has been achieved, many studies \cite{matsunaga2022fine,dhariwal2021diffusion,nichol2021improved} have indicated that GAN-based image translation methods tend to show limited capacity for preserving semantic details.
Unlike existing methods that handle image translation based on the input source image, our method enables the model to translate images further conditioned on the source labels, which enables fine-grained semantic preservation.

\textbf{Denoising Diffusion Probabilistic Models (DDPMs)}, also namely diffusion models, are a class of generative models inspired by the nonequilibrium thermodynamics. 
Diffusion models \cite{sohl2015deep,ho2020denoising} generally learn a denoising model to gradually denoise from an original common distribution, e.g., Gaussian noise, to a specific data distribution.
It is first proposed by Sohl-Dickstein et al. \cite{sohl2015deep}, and has recently attracted much attention in a wide range of research fields including computer vision \cite{dhariwal2021diffusion,vahdat2021score,foo2023diffusion,ackermann2022high,meng2021sdedit,Choi_2021_ICCV,Choi_2021_ICCV,gong2022diffpose,chen2022diffusiondet,gu2022stochastic}, nature language processing \cite{austin2021structured,gong2022diffuseq,li2022diffusion}, audio processing \cite{huang2022prodiff,huang2022prodiff,kim2022guided,tae2021editts,wu2022itowave} and multi-modality \cite{kim2022diffusionclip,liao2023text,huang2023noise2music,voynov2022sketch}. Based on the diffusion technique, Dhariwal et al. proposed a gradient guidance method \cite{dhariwal2021diffusion} to enable pre-trained unconditional diffusion models to generate images based on a specified class label. 
Although effective, the gradient guidance method generally works with image-level labels and is limited to a single domain, making it challenging to be adopted for DASS.
To address these issues, in this paper, we propose a novel Semantic Gradient Guidance (SGG) method and a Progressive Translation Learning (PTL) strategy, enabling the utilization of pixel-level labels for label-guided cross-domain image translation.


\section{Revisiting Diffusion Models}

As our method is proposed based on the Denoising Diffusion Probabilistic Model (DDPM) \cite{sohl2015deep,ho2020denoising}, here, we briefly revisit DDPM. 
DDPM is a generative model that converts data from noise to clean images by gradually denoising.
A standard DDPM generally contains two processes: a diffusion process and a reverse process. 
The diffusion process contains no learnable parameters and it just used to generate the training data for diffusion model's learning.
Specifically, in the diffusion process, a slight amount of Gaussian noise is repeatedly added to the clean image $x_0$, making it gradually corrupted into a noisy image $x_{T}$ after $T$-step noise-adding operations.
A single step of diffusion process (from $x_{t-1}$ to $x_{t}$) is represented as follows: 
\begin{equation}
\label{eq:ddpm_q}
    q(x_t | x_{t-1}) = \mathcal{N}(\sqrt{1-\beta_t}x_{t-1}, \beta_t \textbf{I}),
\end{equation}
where $\beta_t$ is the predefined variance schedule. Given a clean image $x_0$, we can obtain $x_T$ by repeating the above diffusion step from $t=1$ to $t=T$. 
Particularly, an arbitrary image $x_t$ in the diffusion process can also be directly calculated as:
\begin{equation}
\label{eq:ddpm_diffuse}
    x_{t} = \sqrt{\bar{\alpha}_t} x_0 + \sqrt{1-\bar{\alpha}_t} \epsilon,
\end{equation}
where $\alpha_t = 1-\beta_t, \bar{\alpha}_t = \prod_{s=1}^t \alpha_s$ and $\epsilon \sim \mathcal{N}(0,I)$.
Based on this equation, we can generate images from $x_0$ to $x_T$ for training the diffusion model.

The reverse process is opposite to the diffusion process, which aims to train the diffusion model to gradually convert data from the noise $x_T$ to the clean image $x_0$. 
One step of the reverse process (from $x_{t}$ to $x_{t-1}$) can be formulated as: 
\begin{equation}
\label{eq:ddpm_reverse}
    p_\theta(x_{t-1}|x_t) = \mathcal{N}( \mu_\theta(x_t,t), \Sigma_\theta(x_t,t)).
\end{equation}

Given the noisy image $x_T$, we can denoise it to the clean image $x_0$ by repeating the above equation from $t=T$ to $t=1$. In Eq. \ref{eq:ddpm_reverse}, the covariance $\Sigma_\theta$ is generally fixed as a predefined value, while the mean $\mu_\theta$ is estimated by the diffusion model.
Specifically, the diffusion model estimates $\mu_\theta$ via a parameterized noise estimator $\epsilon_\theta(x_t,t)$ which is usually a U-Net \cite{ronneberger2015u}.
Given $x_t$ as the input, the diffusion model will output the value of $\mu_\theta$ as follows:
\begin{equation}
\label{eq:ddpm_mu}
    \mu_\theta(x_t,t) = \frac{1}{\sqrt{1-\beta_t}}\Big(x_t - \frac{\beta_t}{\sqrt{1-\bar{\alpha}_t}}\epsilon_\theta(x_t,t)\Big),
\end{equation}

For training the diffusion model, the images generated in the diffusion process are utilized to train the noise estimator $\epsilon_\theta$ to predict the noise added to the image, which can be formulated as:
\begin{equation}
\label{eq:ddpm_loss}
    L_{DM}(x)=\left \|\epsilon_\theta (x_t,t)- \epsilon \right \| ^2,
\end{equation}
where $\epsilon \sim \mathcal{N}(0, \mathbf{I})$ is the noise added to the image $x_t$.

\section{Method}

We address the problem of domain adaptive semantic segmentation (DASS), where we are provided the source images $\mathcal{X}^S$ with corresponding source labels $\mathcal{Y}^S$ and the target images $\mathcal{X}^T$ (without target labels). Our goal is to train a label-guided image translation model to translate source images into target-like ones. Then we can use the translated data to train a target-domain segmentation model.
To achieve this goal, we first train a basic image translation model based on the diffusion technique, which is the baseline of our approach (Sec. \ref{baseline}). Then, we impose our SGG method to the baseline model, making it conditioned on the pixel-wise source labels (Sec. \ref{FGG}). We further propose a PTL training strategy to enable our SGG to work across a large domain gap (Sec. \ref{TS}). Finally, we detail the training and inference procedures (Sec. \ref{train and infer}).

\begin{figure}[tp]
\centering{}
 \includegraphics[scale=0.46]{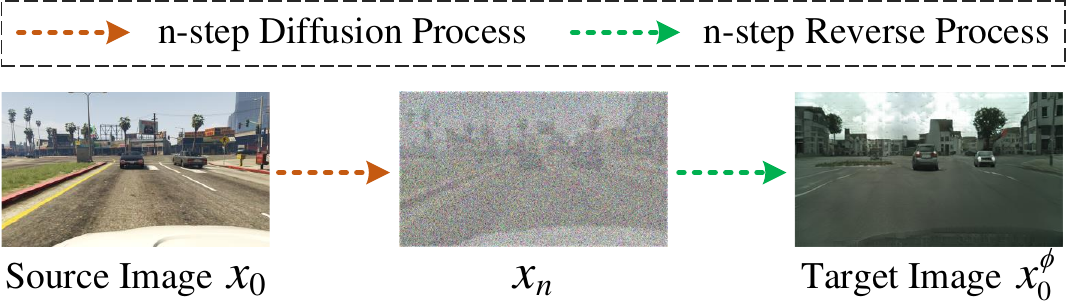} 
 \caption{
Illustration of the Diffusion-based Image Translation Baseline. Starting from a source image $x_0$, the diffusion process disturbs it into a noisy image $x_n$, which is then denoised into the target image $x^{\phi}_0$ using the reverse process performed by a target-pre-trained diffusion model.
}
\label{fig_baseline}
\end{figure}


\subsection{Diffusion-based Image Translation Baseline} \label{baseline}

We start with the baseline of our framework, which is a basic image translation model based on diffusion models.
Before translation, we use the (unlabeled) target domain images $\mathcal{X}^T$ to pre-train a standard diffusion model via Eq. \ref{eq:ddpm_loss}.
As the diffusion model is essentially a denoising model, we use the noise as an intermediary to link the two large-gap domains.
As shown in Fig. \ref{fig_baseline}, we first leverage the diffusion process to add $n$-step noise onto the source image $x_0$ ($x_0 \in \mathcal{X}^S$), making it corrupted into $x_n$, and then use the pre-trained diffusion model to reverse (denoise) it for $n$ steps, obtaining $x^\phi_0$.
Since the diffusion model is pre-trained on the target domain, the reverse (denoising) process will convert
the noise into target-domain content, making the denoised image $x^\phi_0$ shown an appearance closer to the target-domain images than the original source-domain image $x_0$. 
Note that the diffusion model solely trained on the target domain is able to denoise the noisy image from a different source domain towards the target domain, because the diffusion model's training objective is derived from the variational bound on the negative log-likelihood $\mathbb{E}[ -\mathrm{log} p_\theta (\mathcal{X}^T)]$, indicating it learns to generate target-domain data from various distributions.
This property of diffusion model has been verified in \cite{su2022dual} and has been widely used in many studies \cite{meng2021sdedit,gao2022back,Choi_2021_ICCV}.

\begin{figure*}[t]
\begin{centering}
\includegraphics[scale=0.4]{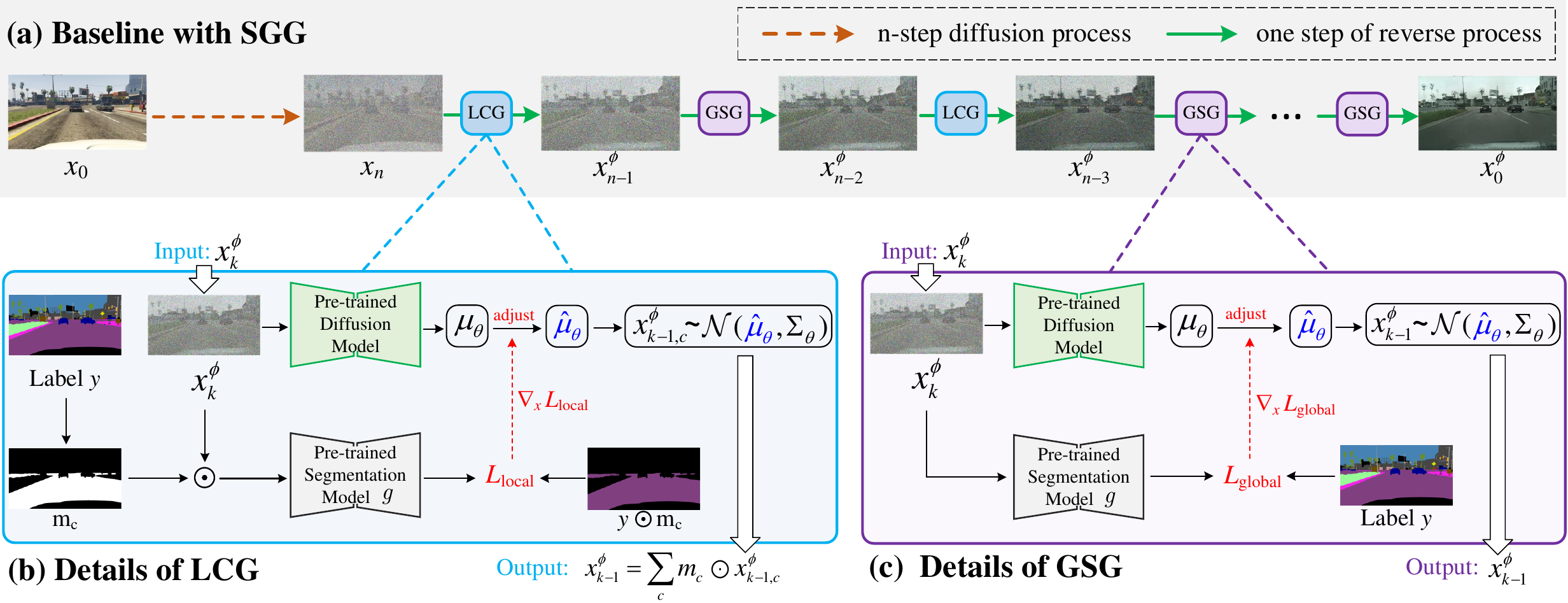} 
\par\end{centering}
\caption{\textbf{(a)} The overall architecture of our baseline model with Semantic Gradient Guidance (SGG). 
SGG alternately adopts the Local Class-regional Guidance (LCG) and the Global Scene-harmonious Guidance (GSG) at each reverse step of the baseline model.
\textbf{(b)} In LCG, we first obtain the loss gradient for class region $m_c$, and then use the gradient to adjust the diffusion model's output $\mu_\theta$ for generating label-related content.
For clarity, we only show the workflow of one class region. 
We obtain the guided results for every class region and finally combine them together via $x^\phi_{k-1}=\sum_{c}m_{c}\odot x^\phi_{k-1,c}$.
\textbf{(c)} In GSG, we calculate the gradient for the whole image and then use the gradient to adjust $\mu_\theta$ for a global-harmony appearance.} \label{fig_fgg} 
\end{figure*}

\subsection{Semantic Gradient Guidance (SGG)}\label{FGG}

The baseline model presented above handles image translation in an unconditional manner, where denoising from much noise however will freely generate new content that may not be strictly consistent with the original image. 
Inspired by the gradient guidance \cite{dhariwal2021diffusion}, we here propose a novel SGG method to enable the unconditional baseline to be conditioned on source labels.

Our SGG consists of two guidance modules, namely the Local Class-regional Guidance (LCG) and the Global Scene-harmonious Guidance (GSG).
They are proposed to affect the reverse process of the baseline model, aiming to guide the denoising process to generate content semantically consistent with source labels.
Specifically, the LCG is proposed to preserve local details, while the GSG is designed to enhance the global harmony of the denoised images.
As shown in Fig. \ref{fig_fgg} (a), we alternately apply the LCG and GSG modules at each reverse step, aiming to provide both local-precision and global-harmony guidance.

\vspace{2mm}
\noindent\textbf{(1) Local Class-regional Guidance (LCG)} 
\vspace{1mm}

The key idea of LCG is to use a pixel classifier to calculate loss to measure if the generated (denoised) pixel has the same semantic as the source semantic label, and then utilize the loss gradient as the guidance signal to guide the pixel to generate same-semantic content.
Assuming that we already have a pre-trained segmentation model $g$ (i.e., the pixel classifier) that can perform pixel classification well over domains (note that we will address this assumption in Sec. \ref{TS}), based on $g$, LCG can carry out guidance as follows. 

Here, we use $x^\phi_k$ to denote the input of the LCG module, and $x^\phi_{k-1}$ as its output, where $k\in \{n,n-2,n-4, ...\}$. 
As shown in Fig. \ref{fig_fgg} (b), given a source label map $y \in \mathcal{Y}^S$, we can leverage $y$ to obtain a binary mask $m_c$ that indicates the region of the $c$-th class in the source image
($c \in \mathcal{C}$ and $\mathcal{C}$ denotes all classes in the source dataset).
Given the input $x^\phi_{k}$, we can multiply $x^\phi_{k}$ with $m_c$ to obtain
the pixels inside the $c$-th class region, which can be fed into the 
pre-trained segmentation model $g$ to predict the label probabilities of the pixels. The output can be used to calculate the cross-entropy loss with 
the source label, aiming to use
the loss gradient to guide the generation of pixels in the $c$-th class region.
As shown in Fig. \ref{fig_fgg} (b), we can use classifier $g$ to compute the cross-entropy loss $ L_{local}$ for pixels in the local $c$-th class region, 
which is formulated as:
\begin{equation}
    L_{local} (x^\phi_{k}, y, c) = L_{ce}\Big(g (x^\phi_{k} \odot m_c), y \odot m_c\Big),
\label{eq:local_loss}
\end{equation}
where $L_{ce}$ is the cross-entropy loss, $\odot$ denotes the element-wise multiplication and $y \odot m_c$ is the source label of the class $c$.

In the standard reverse process, given the noisy image $x^\phi_k$, the diffusion model denoises it by estimating the mean value $\mu_\theta$ (Eq. \ref{eq:ddpm_mu}) and then using it to obtain the denoised image $x^\phi_{k-1}$ via sampling (Eq. \ref{eq:ddpm_reverse}). 
Therefore, the mean $\mu_\theta$ will determine the content of the denoised image.
Based on this, we guide the reverse process by using the computed loss gradient $\nabla_{x^\phi_k} L_{local}$ to adjust $\mu_\theta$, as shown in Fig. \ref{fig_fgg} (b). 
As the loss gradient is computed based on the source label, it can enable the generated content to correspond to the label class. 
We follow the previous gradient guidance method \cite{dhariwal2021diffusion} to formulate the adjustment operation as:
\begin{equation}
    \hat{\mu}_\theta (x^\phi_k, k, c) = \mu_\theta (x^\phi_k, k) + \lambda\Sigma_\theta \nabla_{x^\phi_k} L_{local}(x^\phi_{k}, y, c),
\label{eq:mu}
\end{equation}
where $\mu_\theta$ is the original output of the pre-trained diffusion model , $\Sigma_\theta$ is a fixed value already defined in Eq. \ref{eq:ddpm_reverse} and $\lambda$ is a hyper-parameter which controls the guidance scale. 

Next, we adopt the adjusted mean value $\hat{\mu}_\theta$ to sample the denoised image:
\begin{equation}
   x^\phi_{k-1,c} \sim \mathcal{N}(\hat{\mu}_\theta (x^\phi_k, k, c), \Sigma_\theta).
\label{eq:fake_guided}
\end{equation}

With the above process, the guided denoised image $x^\phi_{k-1,c}$ can preserve the semantics for pixels in the $c$-th class region.
We parallelly perform the above operations for different $c$, and combine different guided regions into a complete image $x^\phi_{k-1}$: 
\begin{equation}
   x^\phi_{k-1}=\sum_{c}m_{c}\odot  x^\phi_{k-1,c},
\label{eq:real_guided}
\end{equation}
where $ x^\phi_{k-1}$ is the output of the LCG module.

\vspace{2mm}
\noindent\textbf{(2) Global Scene-harmonious Guidance (GSG)} 
\vspace{1mm}

The LCG is performed locally, achieving a per-category guidance for image translation. As a complement, we also propose a global guidance module to enhance the global harmony of the translated image, which further boosts the translation quality.

As shown in Fig. \ref{fig_fgg} (c), our proposed GSG also follows the similar operation flow as LCG, i.e., ``computing loss gradient $\rightarrow$ adjusting mean value $\rightarrow$ sampling new image". 
Given $x^\phi_k$ ($k\in \{n-1,n-3,n-5, ...\}$) as the input of the GSG module, the loss function is computed based on the whole image to achieve the global guidance:
\begin{equation}
    L_{global} (x^\phi_{k}, y) = L_{ce}\big( g (x^\phi_{k}), y \big).
\label{eq:global_loss}
\end{equation}

In SGG, we apply the LCG or GSG at each reverse (denoising) step of the baseline model.
By doing so, we can gradually apply the guidance in a step-wise manner, which is suitable for the step-wise translation process of the baseline model.
To achieve both local-precise and global-harmony guidance, we incorporate an equal number of LCG and GSG modules into our approach. Specifically, since the baseline model has $n$ reverse steps, we use $\frac{n}{2}$ LCG modules and $\frac{n}{2}$ GSG modules.
As for the arrangement of LCG and GSG modules, we design several plausible options: (1) first using $\frac{n}{2}$ LCG modules and then $\frac{n}{2}$ GSG modules, (2) first using $\frac{n}{2}$ GSG modules and then $\frac{n}{2}$ LCG modules, (3) alternating between LCG and GSG, and (4) randomly mixing the two types of modules.
We observe that all options perform at the same level. Since the alternating option shows a slightly better performance than others, we adopt the alternating scheme for SGG.

\subsection{Progressive Translation Learning (PTL)}\label{TS}

The proposed SGG method is based on the assumption that we already have a segmentation model trained over domains.
That is, if we want to guide the translation towards the target domain, we need to have a segmentation model that is pre-trained on the target domain.
However, in DASS, we only have labeled images in the source domain for training a source-domain segmentation network.
To enable the SGG to reliably work for translation towards the target domain with the source-domain segmentation model, we propose a Progressive Translation Learning (PTL) strategy.

As mentioned before, our baseline model can directly translate images by first diffusing for $n$ steps and then reversing for $n$ steps (see Fig. \ref{fig_baseline}). We formulate this process as: $\mathcal{X}^n = \phi^n (\mathcal{X}^S)$, where $\mathcal{X}^S$ denotes all source-domain images, $\phi^n$ denotes the baseline model that handles $n$-step diffusion and reverse process, $\mathcal{X}^n$ denotes the translated output images which is related to the parameter $n$.
As the diffusion model can denoise the image towards the target domain, the parameter $n$ essentially controls the amount of target-domain information added to the source image.
Inspired by \cite{gao2022back}, we can flexibly change the domain of translated images by varying the value of $n$.
When $n$ is small, we can obtain translated images $\mathcal{X}^n$ with minor differences to source-domain images $\mathcal{X}^S$.
Conversely, when $n$ is large enough, the translated images $\mathcal{X}^n$ can completely approach the target domain.
By walking through $n$ from 1 to $N$, our baseline model can generate a set of image domains $\{\mathcal{X}^n | n=1,2,...,N\}$, where $\mathcal{X}^N$ approximately follows the target-domain distribution.
Since each step of diffusion/reverse operation only adds/removes a very small amount of noise, the translated images from neighboring domains, i.e., $\mathcal{X}^{n}$ and $\mathcal{X}^{n+1}$, show very slight differences.

Specifically, we start with the source-domain segmentation model $g_0$, which is well-trained on source images $\mathcal{X}^{S}$ with source labels $\mathcal{Y}^{S}$.  
As the domain gap between two adjacent domains is very small, the segmentation network trained on source domain $\mathcal{X}^{S}$ can also work well for the adjacent domain $\mathcal{X}^{1}$.
Based on this property, we can use $g_0$ to carry out SGG guidance for image translation of the next domain $\mathcal{X}^{1}$.
We formulate this operation in a general representation, i.e., given $g_n$ which is well-trained on $\mathcal{X}_n$,
we can guide the translation of $\mathcal{X}_{n+1}$ as follows: 
\begin{equation}
    \mathcal{X}^{n+1} = \mathrm{SGG} (\phi^{n+1} (\mathcal{X}^S) , g_n),
\label{eq:training_1}
\end{equation}
where $\mathrm{SGG} ( \phi^{n+1}, g_n)$ denotes the SGG-guided baseline model with $(n+1)$-step diffusion and reverse process, which is equipped with $g_n$.

Under the $g_0$-based SGG guidance, the translated images $\mathcal{X}^{1}$ have fine-grained semantic consistency with source labels $\mathcal{Y}^S$.
Then, we can use the guided images $\mathcal{X}^{1}$ to fine-tune the segmentation model $g_0$, making it further fit with this domain. 
This operation is also formulated in a general representation, i.e., given the guided images $\mathcal{X}^{n+1}$, we
can fine-tune the segmentation model $g_n$ as follows:
\begin{equation}
L_{ft}=L_{ce}\big(g_n (\mathcal{X}^{n+1} ), \mathcal{Y}^{S} \big)+L_{ce}\big(g_n (\mathcal{X}^{S} ), \mathcal{Y}^{S} \big).
\label{eq:training_2}
\end{equation}
We name the fine-tuned segmentation model as $g_{n+1}$.
The former loss component aims to use the guided images $\mathcal{X}^{n+1}$ to adapt the segmentation model to the new domain.
The latter component is to train the segmentation model on the source domain images $\mathcal{X}^{S}$. We empirically observe  combining both that augments the model’s learning scope leads to slightly better results than using the former loss only.

After fine-tuning the segmentation model $g_0$ on $\mathcal{X}^{1}$, we can perform SGG-guided image translation again using the fine-tuned model (namely $g_1$), obtaining guided images $\mathcal{X}^{2}$. 
Then we can use $\mathcal{X}^{2}$ to further fine-tune the segmentation model $g_1$ for adapting to a new domain.
In this manner, we can perform Eq. \ref{eq:training_1} and Eq. \ref{eq:training_2} iteratively with $n$ increasing from $1$ to $N$, finally obtaining the translated target-domain images $\mathcal{X}^{N}$ and target-fine-tuned
segmentation model $g_N$.
We use the segmentation model $g_N$ for test-time inference. 
Throughout the whole process of PTL, the segmentation model always performs SGG guidance at the adjacent domain with little domain discrepancy, which is the key to enabling the SGG to work across a large domain gap.

\subsection{Training and Inference}\label{train and infer}
\noindent \textbf{Training.} First, we train a standard diffusion model with the unlabeled target-domain images $\mathcal{X}^T$ via Eq. \ref{eq:ddpm_loss}. 
After that, we keep the pre-trained diffusion model frozen and use it to denoise the source-diffused image, which forms our basic image translation model, namely baseline model $\phi^{n}$ (Sec. \ref{baseline}). 
Our baseline model $\phi^{n}$ can translate source images $\mathcal{X}^S$ into the images $\mathcal{X}^n$.
The model $\phi^{n}$ is flexible with a variable parameter $n$, which can control the domain of the translated images $\mathcal{X}^n$. When $n=1$, the translated images are extremely close to the source domain. When $n=N$, the translated images approximately approach the target domain.
We use the source-domain labeled data $\{\mathcal{X}^S,\mathcal{Y}^S\}$ to train a segmentation model $g_0$.
Then, we set $n=1$ for the baseline model $\phi^{n}$ and use it to translate source images $\mathcal{X}^S$ into the adjacent-domain images $\mathcal{X}^1$.
During the translation process, we use the segmentation model $g_0$ to carry out SGG guidance (Eq.\ref{eq:training_1}), obtaining the guided images $\mathcal{X}^1$.
Then, we use the guided images and source labels $\{\mathcal{X}^1, \mathcal{Y}^S\}$ to fine-tune the segmentation model $g_0$ (Eq. \ref{eq:training_2}), obtaining fine-tuned segmentation model $g_1$.
We set $n=2$ for the baseline model $\phi^{n}$, and use $g_1$ to carry out SGG guidance, obtaining the guided images $\mathcal{X}^2$. Next, we can use $\{\mathcal{X}^2, \mathcal{Y}^S\}$ to execute the fine-tuning again.
In this way, we alternate the SGG guidance and the fine-tuning of segmentation model with $n$ growing from 1 to $N$ (i.e., PTL strategy). 
Finally, we obtain the translated target-domain images $\mathcal{X}^N$ and the target-fine-tuned segmentation model $g_N$.
\noindent \textbf{Inference.} After training, we directly use the segmentation model $g_N$ for inference.


\section{Experiment}
\subsection{Implementation Details}\label{Implementation}

\noindent \textbf{Datasets.}
As a common practice in DASS, we evaluate our framework on two standard benchmarks (GTA5 \cite{richter2016playing} $\rightarrow$ Cityscapes \cite{cordts2016cityscapes}, and SYNTHIA \cite{ros2016synthia} $\rightarrow$ Cityscapes \cite{cordts2016cityscapes}).
GTA5 and SYNTHIA provide 24,996 and 9400 labeled images, respectively.
Cityscapes consists of 2975 and 500 images for training and validation respectively.
Following the standard protocol \cite{hoffman2016fcns}, we report the mean intersection over Union (mIoU) on 19 classes for GTA5 $\rightarrow$ Cityscapes and 16 classes for Synthia $\rightarrow$ Cityscapes.

\noindent \textbf{Network Architecture.}
To make a fair and comprehensive comparison, we test three typical types of networks as the segmentation model $g$: 
(1) DeepLab-V2 \cite{chen2017deeplab} architecture with ResNet-101 \cite{he2016deep} backbone; 
(2) FCN-8s \cite{long2015fully} architecture with VGG16 \cite{simonyan2014very} backbone; 
(3) DAFormer \cite{hoyer2022daformer} architecture with SegFormer \cite{xie2021segformer} backbone.
All of them are initialized with the network pre-trained on ImageNet \cite{krizhevsky2017imagenet}.
For the diffusion model, we follow the previous work \cite{ho2020denoising} to adopt a U-Net \cite{ronneberger2015u} architecture with Wide-ResNet \cite{zagoruyko2016wide} backbone.
The diffusion model is trained from scratch without loading any pre-trained parameters.

\noindent \textbf{Parameter Setting.}
Following DDPM \cite{ho2020denoising}, we train the diffusion model with a batch size of 4, using the Adam \cite{kingma2014adam} as the optimizer with learning rate as $2 \times 10^{-5}$ and momentum as 0.9 and 0.99. 
Specifically, for the hyper-parameters of diffusion model, we still follow \cite{ho2020denoising} to set $\beta_t$ from $\beta_1 = 10^{-4}$ to $\beta_T = 0.02$ linearly and $T=1000$. 
The covariance $\Sigma_\theta$ is fixed and defined as $\Sigma_\theta = \beta_t \textbf{I}$.
We train the segmentation model $g$ by using the SGD optimizer \cite{kiefer1952stochastic} with a batch size of 2, a learning rate of $2.5 \times 10^{-4}$ and a momentum of 0.9.  
In SGG, we set $\lambda=80$.
In PTL, we set $N=600$.
We pre-train the segmentation model on the source domain for 20,000 iterations and fine-tune it on other intermediate domains for 300 iterations each. Since the discrepancy between adjacent domains is minor, fine-tuning for 300 iterations is enough to effectively adapt to a new domain.
As the SGG guidance needs the segmentation model $g$ to classify for noisy and masked images, during the training of the segmentation model, we also generate noisy and masked images for data augmentation to train a robust segmentation model.

\begin{table}
\centering
\caption{Performance comparison in terms of mIoU ($\%$). 
Three backbone networks, ResNet-101 (Res), VGG16 (Vgg) and SegFormer (Seg), are used in our study.
The results of the task ``GTA5 $\rightarrow$ Cityscapes" (``SYNTHIA $\rightarrow$ Cityscapes") are averaged over 19 (16) classes. $*$ denotes previous image translation methods.}
\vspace{1mm}
\label{tab:results}
\resizebox{0.45\textwidth}{!}{%
\begin{tabular}{lclccclccc} 
\toprule[0.15em] 
\multicolumn{1}{c}{\multirow{2}{*}{Method}} & \multirow{2}{*}{Venue}      & \multirow{2}{*}{} & \multicolumn{3}{c}{GTA5~→ Cityscapes}         &  & \multicolumn{3}{c}{SYN.~→ Cityscapes}       \\ 
\cline{4-6}\cline{8-10}
\specialrule{0em}{1pt}{1pt}
\multicolumn{1}{c}{}                        &                             &                   & Res             & Vgg             & Seg             &  & Res             & Vgg             & Seg              \\ 
\midrule[0.15em]
\specialrule{0em}{1pt}{1pt}
\specialrule{0em}{1pt}{1pt}
Source only                                 & -                           &                   & 37.3          & 27.1          & 45.4          &  & 33.5          & 22.9          & 40.7           \\
CyCADA \cite{hoffman2018cycada}$*$                                      & ICML'18                     &                   & 42.7          & 35.4          & -             &  & -             & -             & -              \\
IIT \cite{murez2018image}$*$                                      & CVPR'18                     &                   & -          & 35.7          & -             &  & -             & -             & -              \\
CrDoCo \cite{chen2019crdoco}$*$                                       & CVPR'19                     &                   & -             & 38.1          & -             &  & -             & 38.2          & -              \\
BDL \cite{li2019bidirectional}$*$                                         & CVPR'19                     &                   & 48.5          & 41.3          & -             &  & -             & 33.2          & -              \\
CLAN \cite{luo2019taking}                                       & CVPR'19                     &                   & 43.2          & 36.6          & -             &  & -             & -             & -              \\
Intra \cite{pan2020unsupervised}                                      & CVPR'20                     &                   & 45.3          & 34.1          & -             &  & 41.7          & 33.8          & -              \\
SUIT \cite{li2020simplified}$*$                                       & PR'20                     &                   & 45.3          & 40.6          & -             &  & 40.9          & 38.1          & -              \\
LDR  \cite{yang2020label}                                         & ECCV'20                     &                   & 49.5          & 43.6          & -             &  & -             & 41.1          & -              \\
CCM \cite{li2020content}                                        & ECCV'20   &                   & 49.9          & -             & -             &  & 45.2          & -             & -              \\
ProDA \cite{zhang2021prototypical}                                      & CVPR'21                     &                   & 57.5          & -             & -             &  & 55.5          & -             & -              \\
CTF \cite{ma2021coarse}                                        & CVPR'21                     &                   & 56.1          & -             & -             &  & 48.2          & -             & -              \\
CADA \cite{yang2021context}$*$                                       & WACV'21                     &                   & 49.2          & 44.9            & -             &  & -          & 40.8             & -              \\
UPLR \cite{wang2021uncertainty}                                       & ICCV'21                     &                   & 52.6          & -             & -             &  & 48.0          & -             & -              \\
DPL \cite{cheng2021dual}                                        & ICCV'21                     &                   & 53.3          & \underline{46.5}  & -             &  & 47.0          & \underline{43.0}  & -              \\
BAPA \cite{liu2021bapa}                                      & ICCV'21                     &                   & 57.4          & -             & -             &  & 53.3          & -             & -              \\
CIR \cite{gao2021addressing}$*$                                      & AAAI'21                     &                   & 49.1          & -             & -             &  & 43.9          & -             & -              \\
CRAM \cite{zhang2022confidence}                                      & TITS'22                     &                   & 48.6          & 41.8             & -             &  & 45.8          & 38.7             & -              \\
ADAS \cite{lee2022adas}$*$                                      & CVPR'22                     &                   & 45.8          & -             & -             &  & -          & -             & -              \\
CPSL \cite{li2022class}                                       & CVPR'22                     &                   & 55.7          & -             & -             &  & 54.4          & -             & -              \\
FDA \cite{yang2020fda}                                        & CVPR'22                     &                   & 50.5          & 42.2          & -             &  & -             & 40.0          & -              \\
DAP \cite{huo2022domain}                                        & CVPR'22                     &                   & \underline{59.8}          & -             & -             &  & \underline{59.8}  & -             & -              \\
DAFormer \cite{hoyer2022daformer}                                   & CVPR'22                     &                   & -             & -             & 68.3          &  & -             & -             & 60.9           \\
HRDA  \cite{hoyer2022hrda}                                      & ECCV'22                     &                   & -             & -             & \underline{73.8}  &  & -             & -             & \underline{65.8}   \\
ProCA \cite{jiang2022prototypical}                                      & ECCV'22                     &                   & 56.3          & -             & -             &  & 53.0          & -             & -              \\
Bi-CL \cite{lee2022bi}                                      & ECCV'22                     &                   & 57.1          & -             & -             &  & 55.6          & -             & -              \\
Deco-Net \cite{lai2022decouplenet}                                   & ECCV'22                     &                   & 59.1          & -             & -             &  & 57.0          & -             & -              \\
\specialrule{0em}{1pt}{1pt}
\hline
\specialrule{0em}{1pt}{1pt}
Ours                                        & ICCV'23                           &                   & \textbf{61.9} & \textbf{48.1} & \textbf{75.3} &  & \textbf{61.0} & \textbf{44.2} & \textbf{66.5}  \\
\bottomrule[0.15em] 
\end{tabular}}
\end{table}

\begin{figure*}[t]
\begin{centering}
\includegraphics[scale=0.287]{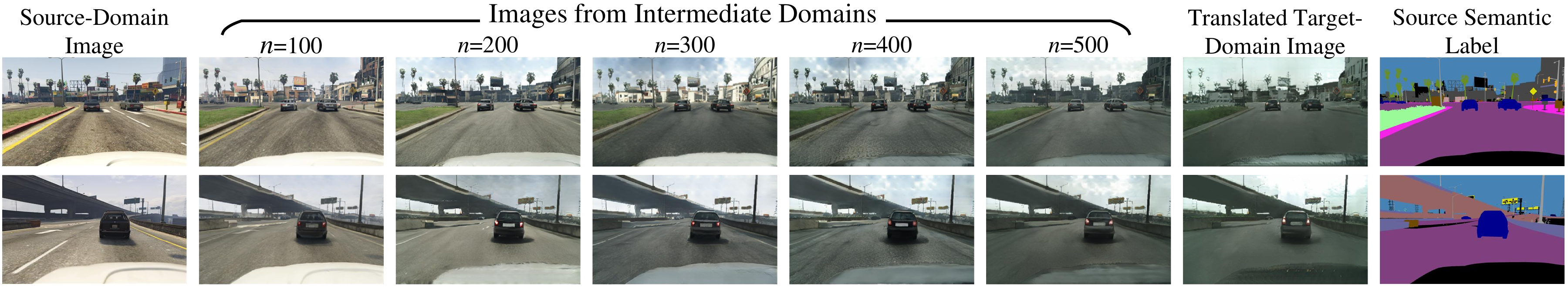} 
\par\end{centering}
\caption{The translated images of our approach (including images from intermediate domains). $n$ denotes the serial number of intermediate domains. We can see that the semantic knowledge is transferred smoothly and precisely.}  \label{fig_domains} 
\vspace{-2mm}
\end{figure*}


\subsection{Comparative Studies}
We compared the results of our method with the state-of-the-art methods on two DASS tasks, i.e., ``GTA5 $\rightarrow$ Cityscapes" and ``SYNTHIA $\rightarrow$ Cityscapes", using three different backbone networks: ResNet-101 \cite{he2016deep}, VGG16 \cite{simonyan2014very} and SegFormer \cite{xie2021segformer}.
To ensure reliability, we run our model five times and report the average accuracy.
The comparative results are reported in Tab. \ref{tab:results}. 
Overall, our method consistently gains the best performance across all settings and backbones. 
Remarkably, when compared to existing image translation methods (marked with $*$), our method brings 3.2\% $\sim$ 20.1\% improvement on all backbones and settings, which demonstrates the superiority of our diffusion-based approach.

\subsection{Ablation Studies}

\begin{table}
\centering
\caption{Ablation studies of different components on the task ``GTA5 $\rightarrow$ Cityscapes" (G$\rightarrow$C) and ``SYNTHIA $\rightarrow$ Cityscapes" (S$\rightarrow$C) with backbone Res-101.}
\vspace{1mm}
\label{tab:ablation}
\resizebox{0.26\textwidth}{!}{
\begin{tabular}{llcc} 
\toprule[0.15em] 
\multicolumn{2}{c}{Model} & G$\rightarrow$C & S$\rightarrow$C   \\ 
\bottomrule[0.10em] 
\specialrule{0em}{1pt}{1pt}
(a)  & Baseline Model & 48.3 & 45.1\\ 
\hdashline[2pt/3pt]
(b)   & Ours w/o SGG   & 55.4 & 52.7\\
(c)   & Ours w/o PTL   & 58.3 & 56.5\\
\hdashline[2pt/3pt]
(d)   & Ours w/o LCG   & 59.8 & 59.1\\
(e)   & Ours w/o GSG   & 60.5 & 59.4\\
\hdashline[2pt/3pt]
(f)   & Ours (full)    & 61.9 & 61.0\\
\bottomrule[0.15em] 
\end{tabular}
\vspace{-4mm}
}

\end{table}

\noindent \textbf{Effect of Baseline model.} 
In Sec. \ref{baseline}, we propose a Diffusion-based Image Translation Baseline, i,e, the baseline model $\phi^n$.
To study the effectiveness of the baseline model $\phi^n$, we set $n=N$ to directly translate source images into the target domain.
After translation, we use the translated images and source images, along with the source labels, to train a segmentation model for inference.
As shown in Tab. \ref{tab:ablation} (a), our baseline model achieves 48.3\% and 45.1\% mIoU on two DASS tasks, respectively. This performance is comparable to the results of previous image translation methods (marked with $*$ in Tab. \ref{tab:results}), which demonstrates the effectiveness of our baseline model.

\noindent \textbf{Effect of SGG.} 
To demonstrate the effectiveness of SGG, during the PTL training, we remove the SGG guidance and directly use the unguided images of each intermediate domain to fine-tune the segmentation model.
By comparing the results in Tab. \ref{tab:ablation} (b) and (f), we can see that when adopting SGG, our framework shows a significant improvement (more than 6\%) on both tasks, clearly demonstrating the effectiveness of SGG in preserving semantics.

\noindent \textbf{Effect of PTL.} 
The key design of PTL is to generate intermediate domains to decompose the large gap into small ones.
To investigate the effectiveness of PTL, we construct an ablation experiment by removing the generation of intermediate domains, i.e., we set $n=N$ for the baseline model $\phi^n$ and use it to translate source images into the target domain, i.e., $\mathcal{X}^N=\phi^N(\mathcal{X}^S)$. During the translation, we directly use the source-trained segmentation model to execute SGG guidance to guide the translation process.
Finally, we use the guided target-domain images to fine-tune the segmentation model.
From Tab. \ref{tab:ablation} (c) and (f), we can see that our framework without PTL strategy consistently incurs an obvious performance drop on both DASS tasks, which demonstrates the effectiveness of PTL strategy.
We show in Fig. \ref{fig_domains} some translated image examples and their corresponding intermediate domains.  
We can see that the image domain is gradually transferred, and all images contain semantics that are well preserved in a fine-grained manner, which demonstrates that our PTL can enable the SGG to reliably work on each intermediate domain.

\begin{figure}[t]
    \centering
    \includegraphics[scale=0.23]{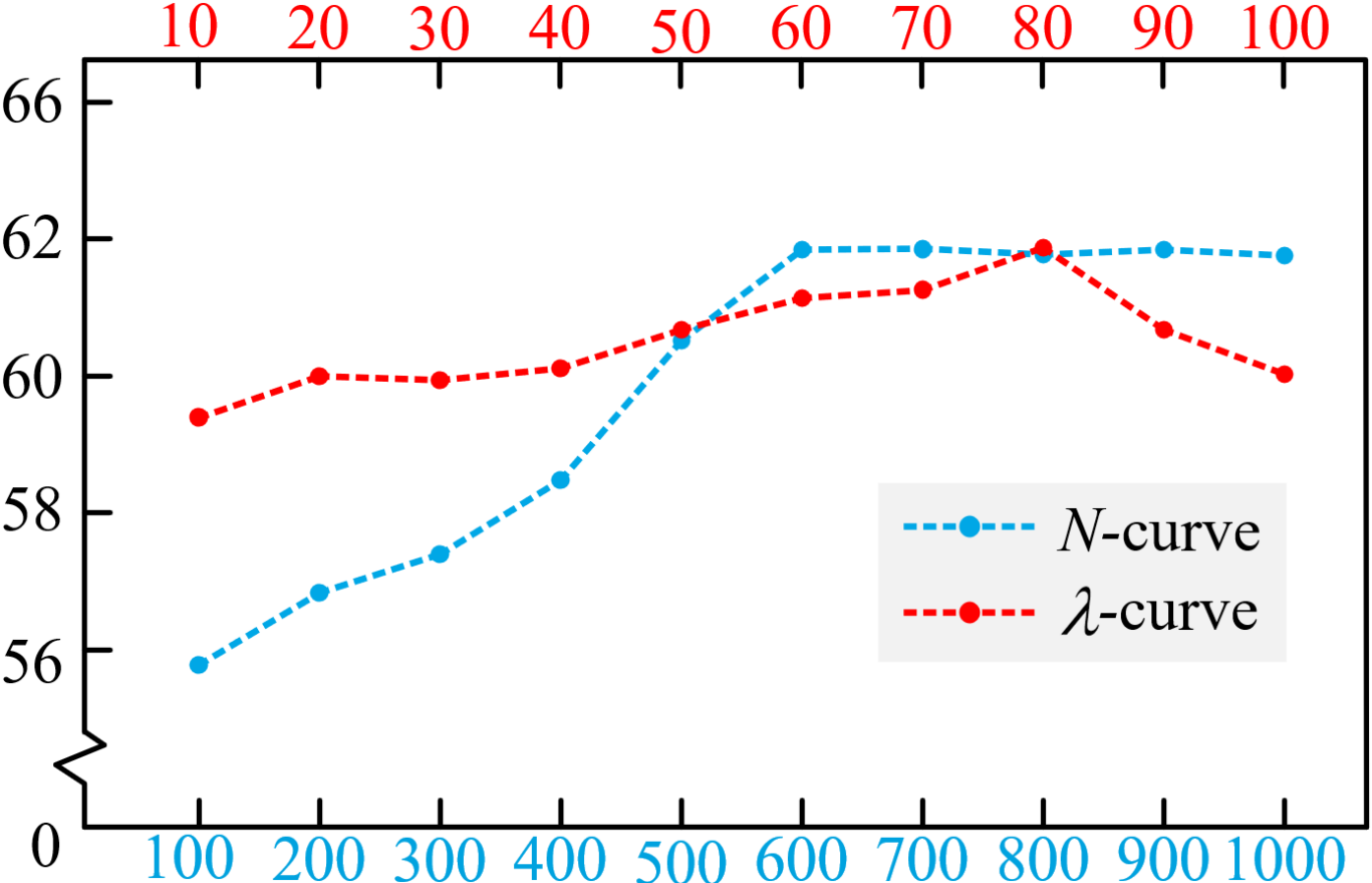}
    \caption{Parameter analysis on $N$ and $\lambda$.}
    \label{fig_parameter}
    \vspace{-2mm}
\end{figure}

\noindent \textbf{Effect of LCG and GSG.} 
In order to evaluate the impact of the LCG and GSG modules, we conduct ablation experiments by removing each module respectively. As shown in Tab. \ref{tab:ablation} (d) and (e), we find that the model’s performance without LCG (or GSG) degrades, which demonstrates their usefulness.
Furthermore, to comprehensively demonstrate the effect of LCG and GSG modules, we provide some qualitative ablation results in \textit{Supplementary}.

    \begin{figure*}[t]
    \begin{centering}
    \includegraphics[scale=0.32]{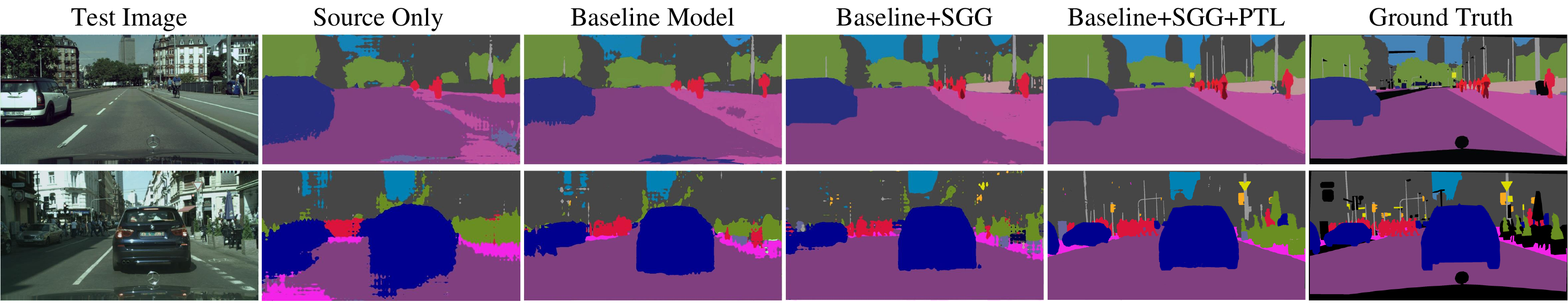} 
    \par\end{centering}
    \vspace{-0mm}
    \caption{Qualitative semantic segmentation results of our framework. It can be seen that the segmentation performance is gradually improved by adding our proposed components one by one. Best viewed in color.} \label{supp-seg} 
    \vspace{-0mm}
    \end{figure*}

\noindent \textbf{Effect of the second loss component of fine-tuning loss in PTL.}
As mentioned in Sec. \ref{TS}, in our PTL strategy, we propose a fine-tuning loss $L_{ft}$ to progressively fine-tune the segmentation model towards the target domain, which is formulated as follows:
    \vspace{-0 mm}
    \begin{equation}
    \label{eq:second_loss}
    L_{ft} = \underbrace{L_{ce}\big(g_n (\mathcal{X}^{n+1} ), \mathcal{Y}^{S} \big)}_{\Large {L_{ada}}} + \underbrace{L_{ce}\big(g_n (\mathcal{X}^{S} ), \mathcal{Y}^{S} \big)}_{ L_{src}}.
    \end{equation}
    \vspace{-0 mm}
The fine-tuning loss has two components. The former loss component $L_{ada}$ aims to adapt the segmentation model to a new domain, which is closer to the target domain. The later component $L_{src}$ is to train the segmentation model on the source domain. 
As shown in Tab. \ref{tab:loss_component}, compared to using the former loss $L_{ada}$ only, the combination of both components can augment the model’s learning scope, leading to better results (i.e., a gain of 0.4\% and 0.3\%).

\begin{table} [h]
\centering
\caption{Ablation study on the loss component $L_{src}$ in fine-tuning loss $L_{ft}$.}
\label{tab:loss_component}
\resizebox{0.25\textwidth}{!}{
\begin{tabular}{llcc} 
\toprule[0.15em] 
\multicolumn{2}{c}{Model} & G$\rightarrow$C & S$\rightarrow$C   \\ 
\bottomrule[0.10em] 
\specialrule{0em}{1pt}{1pt}
(a)  & $L_{ada}$    & 61.5  & 60.7   \\
\specialrule{0em}{1pt}{1pt}
(b)~ & $L_{ada}+L_{src}$   & 61.9  & 61.0   \\
\bottomrule[0.15em] 
\end{tabular}
}
\end{table}

\noindent \textbf{Parameter Analysis.} 
For parameters $\beta$, $T$ and $\Sigma_\theta$, we follow the previous work DDPM \cite{ho2020denoising} to set their values.
In this paper, we only need to study the impact of $N$ (the number of domains in PTL) and $\lambda$ (the guidance scale parameter in SGG).
As for $N$, we desire the number of domains as large as possible, as we want to smoothly bridge the domain gap.
On the other hand, we need to avoid inducing too much training time, i.e., the number of domains should be limited.
We present the ablation results of $N$ in Fig. \ref{fig_parameter} (blue curve). We find the accuracy begins to level off when $N > 600$. We thus set the optimal value of $N$ to 600 with the consideration of saving training time.
As for $\lambda$, we study its impact with different values and show the results in Fig. \ref{fig_parameter} (red curve).
We can see that the best option of $\lambda$ is $\lambda=80$.


\begin{table}[t]
\centering
\caption{Comparison of training and inference time.}
\vspace{1mm}
\label{tab:time}
\resizebox{0.35\textwidth}{!}{
\begin{tabular}{lccc} 
\toprule[0.15em] 
\multicolumn{1}{c}{Methods} & Training Time & Inference Time & Acc.       \\ 
\specialrule{0em}{1pt}{1pt}
\hline
\specialrule{0em}{1pt}{1pt}
BDL \cite{li2019bidirectional}                       & 1.8 days                  & 58.40 ms              &  48.5          \\
ProCA \cite{jiang2022prototypical}          & 1.9 days                  & 57.48 ms              &  56.3          \\
Bi-CL \cite{lee2022bi}                       & 1.7 days                  & 57.91 ms              &  57.1          \\
Deco-Net \cite{lai2022decouplenet}               & 1.9 days                  &  59.67 ms            & 59.1           \\
\hline
Ours                        & 2.3 days                  & 57.68 ms              & \textbf{61.9}  \\
\bottomrule[0.15em] 
\end{tabular}
}
\vspace{-4mm}
\end{table}

\noindent \textbf{Training \& Inference Time.} 
We compare our method with other methods in terms of training and inference time on the task ``GTA5$\rightarrow$Cityscapes". Although our model is trained across $N$ intermediate domains ($N=600$), the training time does not increase much, as we only need to fine-tune the model for several iterations on each intermediate domain (more details in Sec. \ref{Implementation}). As shown in Tab. \ref{tab:time}, compared to recent state-of-the-art models, though our method achieves a significant performance gain, it only requires slightly more training time. 
Since our approach does not change the segmentation model's structure, our method performs inference almost the same as others.

\noindent \textbf{Ablation on different arrangement options of LCG and GSG modules in SGG.}
As shown in Fig. \ref{fig_fgg}, in our framework, the LCG and GSG modules are arranged as a sequence with $n$ steps.
Here, we conduct ablation studies to investigate the impact of different arrangement options.
Tab. \ref{tab:arrange} shows the ablation results, where ``LCG/GSG" denotes first using $\frac{n}{2}$ LCG modules and then $\frac{n}{2}$ GSG modules, ``GSG/LCG" represents first using $\frac{n}{2}$ GSG modules and then $\frac{n}{2}$ LSG modules, ``Alternate" refers to alternating between LCG and GSG, and ``RandMix" represents randomly arranging the two types of modules.
We can see that the alternating option performs best, outperforming others by 0.3\%$\sim$0.6\%.
Therefore, we adopt the alternating option to arrange LCG and GSG modules in the SGG scheme.

\noindent \textbf{Qualitative Segmentation Results.} 
Fig. \ref{supp-seg} shows some qualitative segmentation results. 
The ``Source Only" results were obtained by directly applying the segmentation model trained on the source domain to the target domain.
We can see that even only using our baseline model, the segmentation results are improved obviously (e.g., the road and car).
After adding the SGG scheme to our baseline, the boundary of small objects (e.g., the person) becomes more precise, which demonstrates the SGG's effectiveness in preserving details. 
When further using the PTL strategy, our approach can provide more accurate segmentation results, indicating that the PTL can help SGG to preserve details better.

\begin{table} [t]
\centering
\caption{Ablation studies on different arrangement options of LCG and GSG modules in SGG.}
\label{tab:arrange}
\vspace{2mm}
\resizebox{0.22\textwidth}{!}{
\begin{tabular}{llcc} 
\toprule[0.15em] 
\multicolumn{2}{c}{Model} & G$\rightarrow$C & S$\rightarrow$C   \\ 
\bottomrule[0.10em] 
\specialrule{0em}{1pt}{1pt}
(a)~ & LCG/GSG        & 61.4  & 60.5   \\
(b)  & GSG/LCG        & 61.3  & 60.5   \\
(c)  & Alternate      & \textbf{61.9}  & \textbf{61.0}   \\
(d)  & RandMix        & 61.6  & 60.7   \\
\bottomrule[0.15em] 
\end{tabular}
}
\vspace{-4mm}
\end{table}
\section{Conclusion}

In this paper, we have proposed a novel label-guided image translation framework to tackle DASS.
Our approach leverages the diffusion model and incorporates a Semantic Gradient Guidance (SGG) scheme to guide the image translation based on source labels. 
We have also  proposed a Progressive Translation Learning (PTL) strategy to facilitate our SGG in working across a large domain gap.
Extensive experiments have shown the effectiveness of our method.
\vspace{2mm}

\noindent \textbf{Acknowledgement} This research is supported by the National Research Foundation, Singapore under its AI Singapore Programme (AISG Award No: AISG2-PhD-2022-01-027).

{\small
\bibliographystyle{ieee_fullname}
\bibliography{egbib}
}

\clearpage
\appendix

    \begin{figure}[t]
    \begin{centering}
    \includegraphics[scale=0.095]{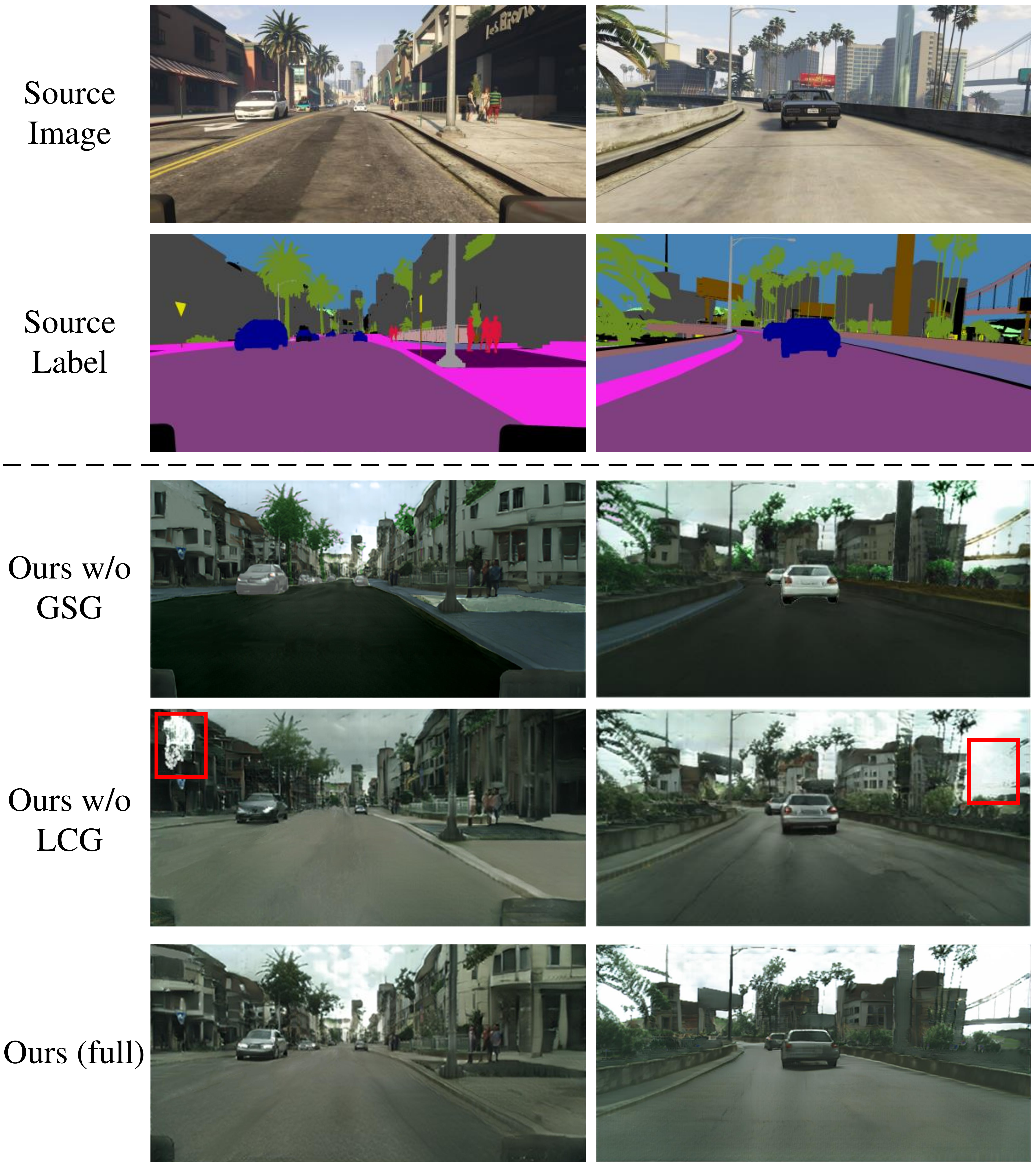} 
    \par\end{centering}
    \vspace{0mm}
    \caption{The visualization of translated images when removing the LCG and GSG modules, respectively.} \label{supp-LCG-GSG} 
    \vspace{-3mm}
    \end{figure}

\section{Effect of LCG and GSG modules}
In the main paper, we already demonstrate the effectiveness of LCG and GSG quantitatively. To provide a more comprehensive ablation study, we present additional qualitative results to visually illustrate the effect of each module.
As shown in the third row of Fig. \ref{supp-LCG-GSG}, the model without GSG produces translated images lacking global harmony, where objects such as cars and trees appear independent from their surroundings, indicating that GSG plays a key role in harmonizing the entire scene.
In contrast, the model without LCG can ensure global harmony but cannot preserve the details well. 
As shown in the fourth row of Fig. \ref{supp-LCG-GSG}, parts of the building and bridge are missing (marked with red boxes).
This observation demonstrates the importance of LCG in preserving local details.
By incorporating both modules, our approach achieves both global-harmony and local-precision image translation, as shown in the last row of Fig. \ref{supp-LCG-GSG}. 
These qualitative results demonstrate the complementary nature of LCG and GSG and show their effects in achieving high-quality image translation results.

    \begin{figure}[t]
    \begin{centering}
    \includegraphics[scale=0.45]{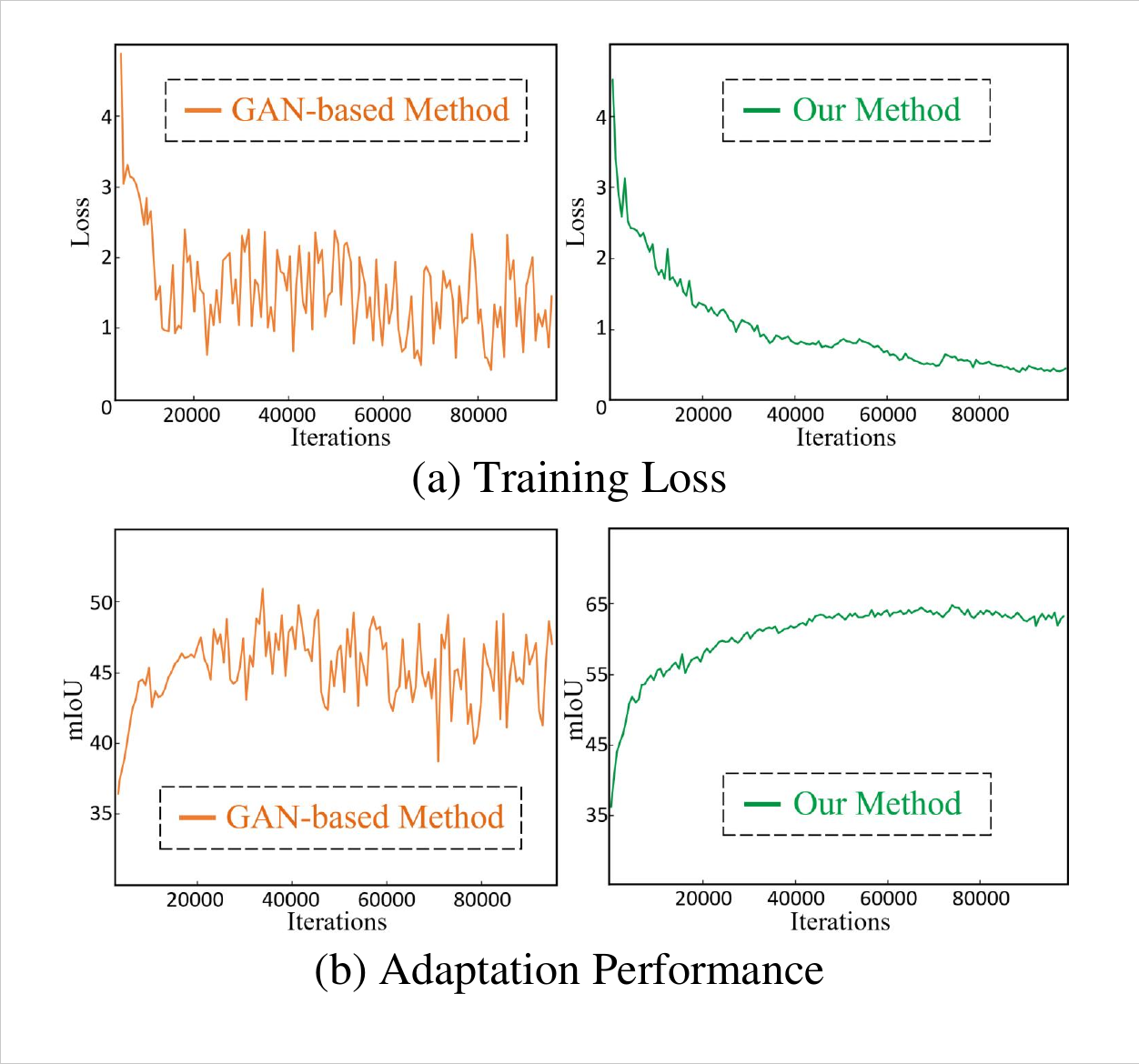} 
    \par\end{centering}
    \vspace{-0mm}
    \caption{The loss and performance curves of the GAN-based image translation method \cite{gao2021addressing} and ours.} \label{supp-training} 
    \vspace{-4mm}
    \end{figure}

\section{Training stability analysis}

In Fig. \ref{supp-training} (a), we show the training loss curves of the state-of-the-art GAN-based image translation method \cite{gao2021addressing} and our diffusion-based method, respectively.
We can observe that our method exhibits a more stable training process.
During the training process, we also use the translated images to train a target-domain segmentation model and then evaluate its adaptation performance on the target domain.
As shown in Fig. \ref{supp-training} (b), we can see that our method achieves a more stable and higher adaptation performance than the GAN-based method, suggesting that our method is not only stable but also effective.

\section{Ablation on data augmentation}

As discussed in Section \textcolor{red}{4.2} of the main paper, our image translation framework is designed to handle noisy and masked images. 
To improve the model's robustness, we generate additional noisy and masked images for data augmentation.
Specifically, we follow the Eqn. \textcolor{red}{2} in the main paper to add noise to the training data,
and then mask the noisy images using binary masks provided by source labels.
We randomly select 10\% of the training data to execute the data augmentation.
As shown in Tab. \ref{tab:augmentation}, the augmentation can bring a slight improvement of 0.2\%$\sim$0.3\%.

\begin{table} [th]
\centering
\caption{Ablation study on data augmentation.}
\label{tab:augmentation}
\vspace{1mm}
\resizebox{0.35\textwidth}{!}{
\begin{tabular}{llcc} 
\toprule[0.15em] 
\multicolumn{2}{c}{Model} & G$\rightarrow$C & S$\rightarrow$C   \\ 
\bottomrule[0.10em] 
\specialrule{0em}{1pt}{1pt}
(a)  & Ours w/o augmentation    & 61.7   & 60.7   \\
(b)  & Ours                     & 61.9   & 61.0   \\
\bottomrule[0.15em] 
\end{tabular}
}
\end{table}

\end{document}